%% file: main.tex
\documentclass{article} 
\usepackage{arxiv}
\date{}
\renewcommand{\undertitle}{}
\renewcommand{\headeright}{}
\usepackage{natbib}
\usepackage{xcolor}
\input{math_commands.tex}

\definecolor{cvprblue}{rgb}{0.21,0.49,0.74}
\definecolor{lightcarminepink}{rgb}{0.9, 0.4, 0.38}
\usepackage[pagebackref,breaklinks,colorlinks,citecolor=gray,linkcolor=lightcarminepink]{hyperref}
\usepackage{url}

\usepackage{booktabs}       
\usepackage{amsfonts}       
\usepackage{nicefrac}       
\usepackage{microtype}      
\usepackage{xcolor}         
\usepackage{array}
\usepackage{graphicx}
\usepackage{orcidlink}
\usepackage{algorithm}
\usepackage{algorithmicx}
\usepackage{algpseudocode}
\usepackage{longtable}
\usepackage{algcompatible}
\usepackage[dvipsnames]{xcolor}
\usepackage{multirow}
\usepackage{graphicx}
\usepackage{subcaption}
\usepackage{mdframed}
\usepackage{wrapfig}
\usepackage{float}
\usepackage{enumitem}
\usepackage{xcolor}
\usepackage[table]{xcolor}
\usepackage{colortbl}
\newcommand{\graycell}[1]{\textcolor{black!55}{#1}}

\title{Branch-Centric Tokenization and Test-Time Augmentation for Skeleton Generation}

\author{Zhengyuan Li \\
Purdue University\\
\And
Chuanyu Pan \\
Meshy AI \\
\And
Yuanming Hu \\
Meshy AI \\
\And
Raymond A. Yeh \\
Purdue University \\
}

\begin{document}

\maketitle
\pagestyle{plain}
\thispagestyle{plain}

\input{figures/fig_teaser}
\begin{abstract}
\input{sections/abs}
\end{abstract}

\input{sections/intro}
\input{sections/rel}

\input{sections/prelim}

\input{sections/method}
\input{sections/exp}
\input{sections/conclusion}

\bibliography{reference}
\bibliographystyle{reference}

\appendix
\input{sections/supp}

\end{document}

%% file: math_commands.tex
\usepackage{amssymb}
\usepackage{amsmath,amsfonts,bm}

\newcommand{\myparagraph}[1]{\vspace*{2pt}{\noindent \bf #1}}

\usepackage{xspace}
\def\@onedot{\ifx\@let@token.\else.\null\fi\xspace}
\DeclareRobustCommand\onedot{\futurelet\@let@token\@onedot}
\def\sp{space}

\newcommand{\figref}[1]{Fig\onedot~\ref{#1}}
\newcommand{\Figref}[1]{Fig\onedot~\ref{#1}}
\newcommand{\equref}[1]{Eq\onedot~\eqref{#1}}
\newcommand{\secref}[1]{Sec\onedot~\ref{#1}}

\def\1{\bm{1}}

\DeclareMathAlphabet{\mathsfit}{\encodingdefault}{\sfdefault}{m}{sl}
\SetMathAlphabet{\mathsfit}{bold}{\encodingdefault}{\sfdefault}{bx}{n}

\def\gC{{\mathcal{C}}}

\def\gE{{\mathcal{E}}}

\def\gJ{{\mathcal{J}}}

\def\gM{{\mathcal{M}}}

\def\gT{{\mathcal{T}}}

\def\gV{{\mathcal{V}}}

\DeclareMathOperator*{\argmin}{arg\,min}

\newcommand{\beas}{\begin{eqnarray*}}
\newcommand{\eeas}{\end{eqnarray*}}
\newcommand{\bea}{\begin{eqnarray}}
\newcommand{\eea}{\end{eqnarray}}
\newcommand{\bes}{\begin{equation*}}
\newcommand{\ees}{\end{equation*}}
\newcommand{\be}{\begin{equation}}
\newcommand{\ee}{\end{equation}}
\makeatletter
\usepackage{xspace}
\def\@onedot{\ifx\@let@token.\else.\null\fi\xspace}
\DeclareRobustCommand\onedot{\futurelet\@let@token\@onedot}
\def\sp{space}

\newcommand{\tabref}[1]{Tab\onedot~\ref{#1}}

\newcommand{\vsp}{\csname v\sp \endcsname}
\newcommand{\hsp}{\csname h\sp \endcsname}

\def\eg{\emph{e.g}\onedot} 
\def\ie{\emph{i.e}\onedot}



%% file: figures/fig_teaser.tex
\begin{figure}[h]
    \vspace{-1.cm}
    \centering
    \includegraphics[width=\linewidth]{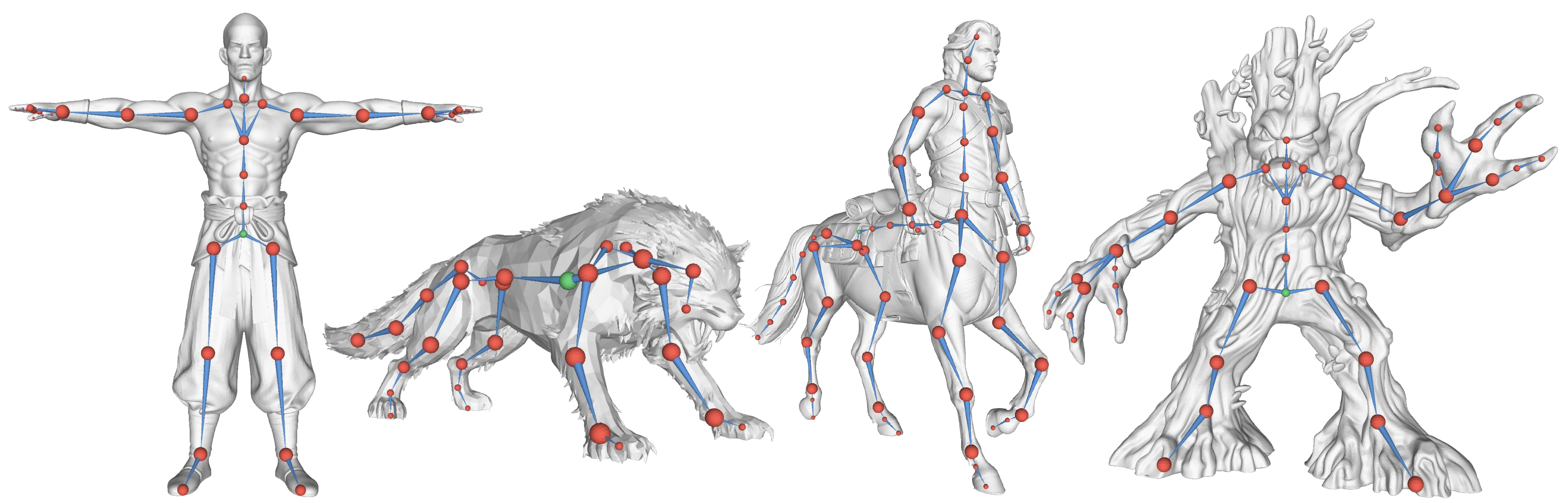}
    \caption{
    Our method predicts skeletons accurately from in-the-wild meshes, including humans, animals, and non-existent fantasy creatures.
    }
    \label{fig:teaser}
    \vspace{-.05cm}
\end{figure}

%% file: sections/abs.tex
Automatic skeleton generation involves predicting both joint positions and skeletal connectivity. However, existing approaches struggle to encode branch structures into token sequences and do not use test-time computation effectively. We study these choices within a unified autoregressive framework. First, we introduce branch-centric tokenization, a branch-aware representation that places structurally related elements next to each other and encodes connectivity directly in the sequence. Compared with standard BFS-style serialization, this representation yields more compact sequences. Second, we introduce view-augmented generation, a test-time augmentation procedure that applies axis-aligned rotations to the input mesh, maps all predictions back to a common frame, and selects the final skeleton based on mesh coverage and consistency among predictions from different views. Experiments show that our method achieves better skeleton prediction accuracy than state-of-the-art methods. In particular, our method reduces the CD-J2B error by $16.9\%$ on the Articulation-XL2.0 dataset compared to the strongest directly comparable baseline, Auto-Connect. Qualitative results on in-the-wild meshes further demonstrate generalization across diverse inputs. 

%% file: sections/intro.tex
\section{Introduction}
\label{sec:intro}
Creating the skeleton of a given static mesh is a crucial step for character animation. 
Traditionally, this process is done manually by skilled rigging artists with domain knowledge of human and animal anatomy. 
With the advancement in deep learning~\citep{vaswani2017attention,brown2020language} and the availability of large rigged-mesh datasets~\citep{xu2019predicting, song2025puppeteer,xu2020rignet,deng2025anymate}, automatic skeleton generation~\citep{guo2025auto,zhang2025one,song2025puppeteer,song2025magicarticulate,liu2025riganything,zhang2026skin,deng2025anymate} has substantially improved. Recent state-of-the-art (SOTA) skeleton generation methods~\citep{zhang2025one,song2025puppeteer,guo2025auto} are inspired by language models~\citep{touvron2023llama,brown2020language}. The main idea is to represent a skeleton as a sequence of tokens 
and autoregressively decode the token sequence with Transformers~\citep{zhang2022opt} given an encoded feature vector of the input mesh~\citep{zhao2023michelangelo,yu2021pointbert}. Despite this progress, some design choices could still be improved. 

First, the quality of these autoregressive models heavily depends on how the skeleton is tokenized. Auto-Connect~\citep{guo2025auto} introduced a unified vocabulary that encodes both joint positions and connectivity. This enables standard autoregressive generation without a separate connectivity module. However, a standard breadth-first search (BFS) traversal is used to generate the sequence from the skeleton, which treats all nodes identically. In other words, the tokenization does not explicitly capture the different structural roles of nodes in the skeleton. Second, existing works~\citep{zhang2025one,zhang2026skin} use \textit{generic} test-time scaling methods to improve performance at the cost of computation, \eg, increasing the beam width during decoding. In this work, we address these two shortcomings. 

Intuitively, a good tokenization should compactly represent both joint positions and connectivity by keeping structurally related joints close together in the token sequence. Classical work on skeleton representation distinguishes junction nodes (degree $> 2$), leaf nodes (degree $= 1$), and degree-2 connection nodes that form branches between them~\citep{cornea2024curve,felzenszwalb2005pictorial,torsello2006learning,siddiqi1999shock,bai2009active,bai2008path}. Inspired by these works, we propose \textbf{branch-centric tokenization}, where the serialization first visits the junction and leaf nodes to identify the branch structure. Then, the connection nodes are placed back between the nodes representing the two ends of a branch. In other words, tokens from the same branch are placed adjacent in the sequence. 

Next, we introduce a test-time augmentation (TTA) approach to improve test-time performance that is complementary to the existing generic test-time scaling. The proposed TTA approach is based on the observation that meshes in existing datasets, \eg, Articulation-XL2.0~\citep{song2025puppeteer}, do not have a consistent upright direction/axis~\citep{zhao2026stroke3d}. Commonly, 3D assets from different sources may appear in different axis-aligned orientations. To address this orientation issue, we perform \textit{view augmentation} on the mesh to generate one skeleton for each rotated view. We then introduce a \textit{coverage-consensus aggregation} to select the final generation based on the coverage of the mesh geometry and agreement among the other predictions.

Empirically, we train a standard autoregressive next-token prediction model~\citep{zhang2022opt} for skeleton generation and incorporate these two improvements; other design choices are kept the same as the baseline SOTA of Auto-Connect. Across Articulation-XL2.0, ModelsResource, and the more challenging diverse-pose split, our model achieves the best results on all reported metrics. Specifically, relative to Auto-Connect on Articulation-XL2.0, our model reduces CD-J2J, CD-J2B, and CD-B2B by $8.1\%$, $16.9\%$, and $14.0\%$, respectively.
%
{\bf Our contributions are as follows:}
\vspace{-3pt}
\begin{itemize}[topsep=0pt, leftmargin=12pt]
    \item We propose branch-centric tokenization, a branch-aware skeleton representation that encodes connectivity directly and places structurally related elements close together in the token sequence. 
    \item We introduce view-augmented generation, a test-time augmentation strategy for skeleton generation that applies axis-aligned input rotations and selects the final prediction based on mesh coverage and cross-view consistency.
    \item Extensive experiments show that our method achieves state-of-the-art skeleton prediction accuracy on both in-dataset and cross-dataset evaluations, and generalizes well to in-the-wild meshes.
\end{itemize}

%% file: sections/rel.tex
\section{Related Work}
\label{sec:rela}
{\bf\noindent Automatic skeleton generation} has made progress from geometric and topology-driven methods to data-driven template-based and template-free models. Early systems embedded a predefined skeleton into a mesh with geometric heuristics~\citep{baran2007automatic}, while curve-skeleton and medial-axis methods extracted topology-aware structures from point clouds or volumetric shape representations~\citep{tagliasacchi2009curve,yan2016erosion,yan2018voxel}. Later template-based methods improved robustness for constrained domains such as humanoids by fitting, transferring, or deforming a predefined rig and learning articulation priors from data~\citep{feng2015avatar,li2021learning,ma2023tarig,chu2025humanrig}.

In contrast, template-free approaches predict joints and connectivity directly from geometry, using volumetric networks, graph neural networks, or motion-aware priors to handle more diverse categories and topologies~\citep{xu2019predicting,xu2020rignet,xu2022morig}. Other recent works~\citep{guo2025make,sun2025drive} go beyond fixed templates but are only demonstrated on human characters, which limits their applicability to more diverse categories.

The availability of large-scale rigging datasets, such as Anymate, Articulation-XL2.0, and Rig-XL~\citep{deng2025anymate,song2025puppeteer,zhang2025one}, has further accelerated recent progress in automatic skeleton generation. More recent methods increasingly cast skeleton generation as structured prediction over a serialized tree. A common pattern is to encode the input shape and autoregressively decode a variable-length skeleton sequence, which better handles varying numbers of joints and bones and their dependencies across categories~\citep{liu2025riganything,song2025magicarticulate,zhang2025one,song2025puppeteer,guo2025auto,sun2026animator,qin2026vip}.

Within this trend, methods still differ in how they represent and predict connectivity. MagicArticulate~\citep{song2025magicarticulate} studies spatial and hierarchical orderings for serialized bone sequences. UniRig~\citep{zhang2025one} and TokenRig~\citep{zhang2026skin} decompose the skeleton into branch-like subsequences, autoregressively predict quantized coordinates for each branch, and reconstruct cross-branch parent-child relations by proximity during detokenization. Anymate~\citep{deng2025anymate} predicts connectivity in a separate stage after joint prediction. RigAnything~\citep{liu2025riganything} and Puppeteer~\citep{song2025puppeteer} both use autoregressive joint sequences with order-dependent parent indices, which couple the representation of a joint to the chosen traversal.

Most related to this work is Auto-Connect~\citep{guo2025auto}, which uses a unified connectivity-preserving tokenization to model topology and coordinates in a single vocabulary, and further improves topological accuracy with a DPO-based post-training stage. Our method is similar to Auto-Connect in model architecture but with a different tokenization design. Also, unlike Auto-Connect, our method does not use post-training and instead studies test-time augmentation.

\myparagraph{Test-Time Augmentation (TTA) and scaling.}
TTA improves prediction by aggregating outputs from multiple transformed versions of the same test input. In image classification, early TTA methods used heuristic multi-crop, flip, and multi-scale evaluation to improve accuracy~\citep{krizhevsky2012imagenet,szegedy2015going,simonyan2014very}. Later work made TTA more adaptive by learning which test-time views to apply or how to aggregate them. For example,~\citet{lyzhov2020greedy} learn a test-time augmentation policy,~\citet{kim2020learning} predict useful transformations for each test instance, and~\citet{shanmugam2021better} replace uniform averaging with learned aggregation weights. However, these methods focus on image classification and do not directly transfer to structured skeleton generation, where predictions may differ in both topology and correspondence; it remains unclear how to aggregate these predictions. Note that, in prior skeleton generation work, test-time scaling mainly relies on beam search for autoregressive decoding~\citep{wu2016google,zhang2025one,zhang2026skin}. In contrast, our method adopts TTA for test-time scaling by generating multiple hypotheses from transformed inputs and aggregating them with a task-specific strategy.

%% file: sections/prelim.tex
\section{Background}
\label{sec:background}
To establish a common notation, we review the task of skeleton generation and the prior work, Auto-Connect~\citep{guo2025auto}. We then provide an overview of the TTA framework. 

{\bf\noindent Task formulation.} Given an input mesh $\gM$, the goal of skeleton generation is to build a model $f_\theta$ that generates the corresponding skeleton $\gT=(\gV,\gE)$ represented as a rooted tree. Here, $\gV$ is the set of nodes denoting the skeleton joints, and $\gE$ is the set of edges between two nodes indicating the parent-child relationship in a tree~\citep{zhang2025one}. Each skeleton joint $v\in\gV$ corresponds to a 3D coordinate of $j^v=(j_x^v, j_y^v, j_z^v)$ in the normalized unit cube space of $[-1,1]^3$. 

Recent works~\citep{guo2025auto} formulate the model $f_\theta$ as a sequence generation task using Transformers~\citep{vaswani2017attention} parametrized by $\theta$. The main innovation is how to convert the skeleton (a tree) into a sequence of tokens $
\boldsymbol{\tau} = (\tau_1, \dots \tau_T),
$
which we now describe in detail.

\input{figures/fig_pipeline}
{\bf\noindent Connectivity-preserving tokenization.}
The main challenge to represent a skeleton $\gT=(\gV,\gE)$ as tokens is to preserve the connectivity information of the tree structure. Tokens should not only encode the coordinates of the joints $v$, but also the parent-child relationships between joints.
To do so, Auto-Connect~\citep{guo2025auto} discretizes the joint coordinates using $n$-bit quantization, where each coordinate is mapped into a discrete index, \ie,
\begin{align}
    Q_n(a) = &\left\lfloor (a+1)\cdot 2^{n-1} \right\rfloor.
\end{align}
Recall that each joint $v$ lies within the normalized cube space $[-1,1]^3$, so the quantization maps each coordinate into an integer index in $\gC_{\tt coord} \triangleq \{0, 1, \dots, 2^n-1\}$. 

To capture the tree structure, a set of four special structure tokens $\gC_{\tt struct}$ is introduced to encode parent-child relationships: [BOS] (beginning of sequence), [EOS] (end of sequence), [E1] (end of a child sequence), and [E2] (end of a level), \ie,
\bea
    \mathcal{C}_{\text{struct}} \triangleq \{ \text{[BOS]}, \text{[EOS]}, \text{[E1]}, \text{[E2]} \}.
\eea
Auto-Connect~\citep{guo2025auto} then determines the ordering of these tokens by traversing the skeleton in a Breadth-First Search (BFS) order, processing the tree level by level.

With the sequence of tokens defined, a Transformer model $f_\theta$ is trained to predict the token sequence $\widehat{\boldsymbol{\tau}} = f_\theta(\gM)$ given the input mesh $\gM$. We refer readers to the original Auto-Connect paper~\citep{guo2025auto} for further details on the training procedure, \eg, losses and DPO post-training (which we do not need).

{\bf\noindent Test-Time Augmentation (TTA).} As the name suggests, TTA is a family of techniques that augment a test input and aggregate the model's predictions across the augmented variants to improve performance. More formally, let $x$ be the test input, $\mathcal{A} = \{a_1, a_2, \dots, a_K\}$ denote a set of transformations, and $f_\theta$ be the model. TTA outputs an aggregated prediction $y^\star$ by applying the model to each transformed input $a_k(x)$ and aggregating the predictions with an aggregation function $g$. In theory, any function that is invariant to the permutation in the index $k$ could work. For tasks such as image classification, a common choice for $g$ is the average function, \ie,
\bea
y^\star = \frac{1}{K}\sum_{k=1}^K f_\theta(a_k(x)).
\eea
In practice, the choice of the transformation set $\mathcal{A}$ and the aggregation function $g$ are defined based on the specific task at hand.

%% file: figures/fig_pipeline.tex
\begin{figure}[t]
    \centering
    \includegraphics[width=0.97\linewidth]{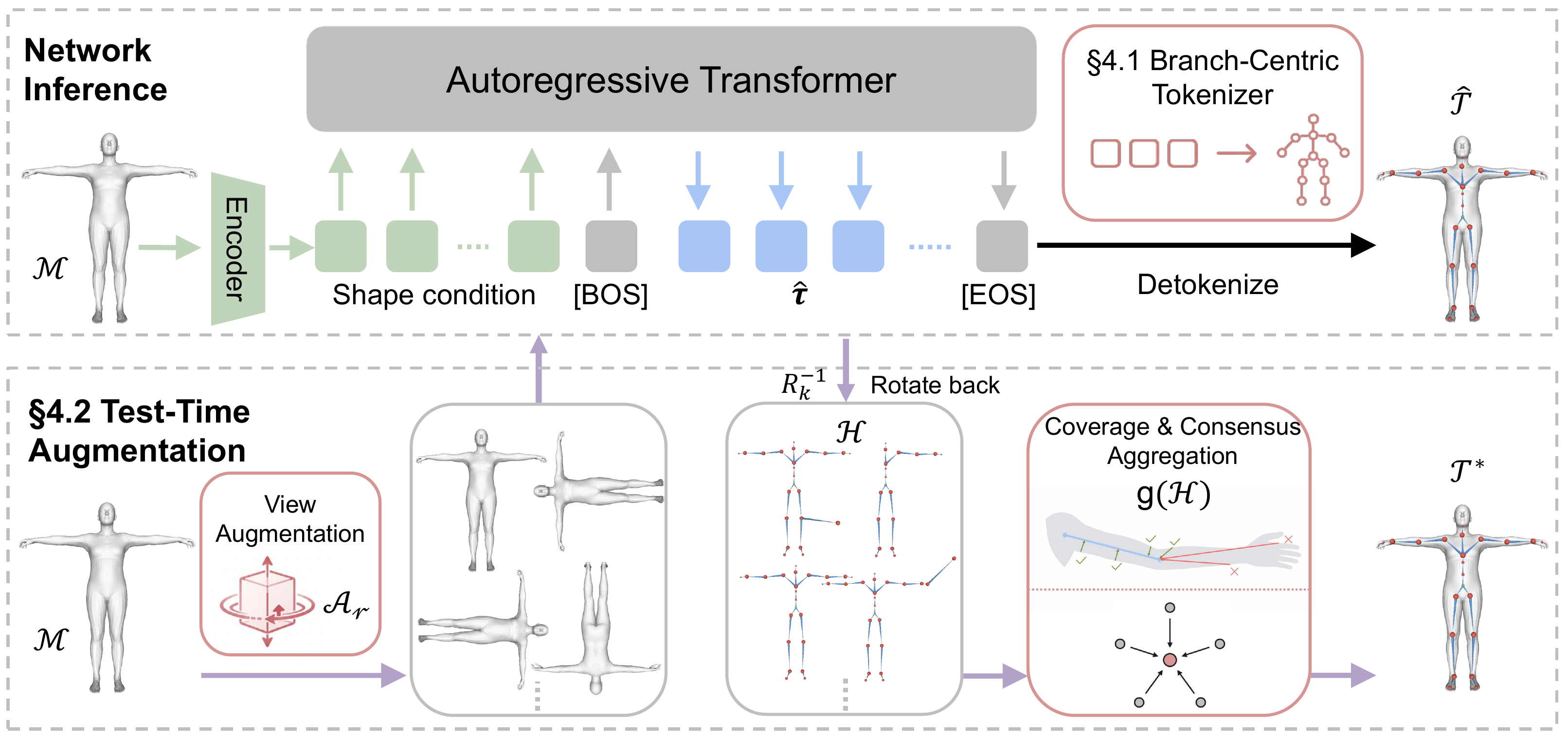}
    \caption{Overview of our method. \textbf{Top: network inference with branch-centric tokenization.} Following the standard mesh-conditioned autoregressive formulation reviewed in~\secref{sec:background}, the input mesh $\gM$ is first encoded into shape-conditioned features, and the autoregressive transformer predicts a token sequence from \texttt{[BOS]} to \texttt{[EOS]}. In~\secref{sec:cct}, we design branch-centric tokenization (BCT), which organizes the skeleton by branch structure and is used to detokenize the predicted sequence $\widehat{\boldsymbol{\tau}}$ into the skeleton tree $\widehat{\gT}$. \textbf{Bottom: test-time augmentation.} In~\secref{sec:reranking}, we apply the rotation set $\mathcal{A}_r$ to the input mesh, decode one skeleton hypothesis for each rotated view, map all predictions back to the canonical frame with the inverse rotations $R_k^{-1}$, and form the hypothesis pool $\mathcal{H}$. Coverage-consensus aggregation $g(\mathcal{H})$ then selects the final prediction $\gT^*$ by favoring hypotheses that both better cover the mesh geometry and agree with the other candidates.
    }
    \label{fig:pipeline}
    \vspace{-0.33cm}
\end{figure}

%% file: sections/method.tex
\section{Method}
\label{sec:method}
We build upon Auto-Connect~\citep{guo2025auto} and improve it in two directions: {\bf (a)} Auto-Connect relies on BFS-based connectivity-preserving tokenization, as reviewed in~\secref{sec:background}. However, this tokenization treats all nodes equally during its traversal, which does not utilize additional node properties that could be beneficial in representing tree topology;  {\bf (b)} Auto-Connect handles the ambiguity of $z$-axis alignment only at training time through random rotation augmentation, which may be insufficient to address the diverse axis-aligned orientations in the data. 
In this work, we propose \emph{branch-centric tokenization} (\secref{sec:cct}), which focuses on viewing a skeleton as a sequence of branches (a.k.a. chains in the graph literature) rather than a sequence of individual joints.
We then introduce a \emph{test-time augmentation} (\secref{sec:reranking}) that augments along different axis alignments and then aggregates to make a final prediction. The overall pipeline is illustrated in~\figref{fig:pipeline}.

\input{figures/fig_tokenization}
\subsection{Branch-Centric Tokenization (BCT)}
\label{sec:cct}
Our proposed tokenization is inspired by classic works on skeleton graphs~\citep{bai2008path,bai2009active}, which define several different node types: a \textit{junction node} is a node with more than two adjacent nodes, a \textit{connection node} is a node with two adjacent nodes, and a \textit{leaf (end) node} is a node with only one adjacent node. A {\it skeleton branch} (series/chain) is defined as a path connecting two non-connection nodes, with all intermediate nodes, if any, being connection nodes. In a skeleton, junction and leaf nodes determine the branching structure, whereas connection nodes mainly describe the within-branch structure.

Existing works using a BFS traversal on the tree do not consider the branching structure, 
and {\bf may split nodes from the same branch into separate parts} of the traversal sequence. See illustration in \figref{fig:token} (right). This ordering requires the model to remember branch information throughout a longer sequence, which we view to be an undesirable property of the representation.

To address this issue, we propose branch-centric tokenization (BCT), which focuses on ensuring the nodes within a branch remain in a chain after the tokenization. As illustrated in~\figref{fig:token}, BCT involves four stages: (a) tree smoothing, (b) BFS on the series-reduced tree, (c) connection node insertion, and (d) delimiter insertion.


{\bf Tree smoothing.} To build a branch-centric representation, we first need to identify all the non-connection nodes. This is done by ``smoothing'' the tree, \ie, replacing the two edges incident to a connection node with a single edge joining its two neighbors and removing that node. We smooth the skeleton tree repeatedly until no vertices of degree two remain, to obtain the series-reduced tree~\citep{gross2018graph} with vertices
\bea
\label{eq:vback}
\mathcal{V}_{\tt sr} \triangleq \{v \in \mathcal{V} \mid \deg(v) \neq 2\} \cup \{v^{\mathrm{root}}\}.
\eea
Here, we always include the root in the set of nodes. This yields the rooted series-reduced tree $\mathcal{T}_{\tt sr}=(\mathcal{V}_{\tt sr},\mathcal{E}_{\tt sr})$, whose edges $\mathcal{E}_{\tt sr}$ correspond to \textit{branches} in the original skeleton tree.

\input{figures/fig_tokencount}%

{\bf BFS on series-reduced tree and connection node insertion.}
We serialize $\mathcal{T}_{\tt sr}$ in breadth-first order starting from $v^{\mathrm{root}}$, which determines an ordering of branches. To obtain a deterministic sibling order, for the children of each series-reduced node $p$, we use a \emph{distance-aware traversal (DAT)} rule and sort them in ascending order of Euclidean distance to $p$, instead of using random shuffling or spatial sequence ordering~\citep{song2025magicarticulate}. This rotation-invariant rule yields a more consistent serialization order under rotations. For each visited series-reduced edge $(p,v)$, we first emit the node $v$ and then insert the intermediate connection nodes on the corresponding branch in the original tree along the path from $v$ back to $p$, yielding the ordered subsequence $\mathcal{S}_{v \to p}$. Each branch forms a contiguous subsequence in the final token sequence.

{\bf Delimiter insertion.} Lastly,  we use the same \texttt{[E1]} and \texttt{[E2]}, as reviewed in~\secref{sec:background},  to mark the end of a child sequence and the end of a BFS level. 
We additionally introduce \texttt{[E3]}  to terminate 
each local branch after all nodes in $\mathcal{S}_{v \to p}$ have been emitted. This extra delimiter makes branch boundaries explicit in the token sequence and helps distinguish starting a new branch from continuing the current one. We provide the full algorithm in Appendix~\secref{sec:bct_algo}.\vspace{4pt}\\

{\bf Benefits of BCT.} Besides the fact that nodes within a branch maintain their closeness in the tokenized sequence, we also observe that BCT leads to a more compact representation. \figref{fig:tokencount} shows the empirical distribution of sequence lengths for our tokenization and Auto-Connect’s tokenization. We observe that our tokenization produces shorter sequences for the large majority of samples, which should be beneficial, as shorter sequences are typically easier to model~\citep{khandelwal2018sharp,press2021shortformer,li2022stability}.


\input{sections/tta.tex}

%% file: figures/fig_tokenization.tex
\begin{figure}[t]
    \centering
    \includegraphics[width=0.98\linewidth]{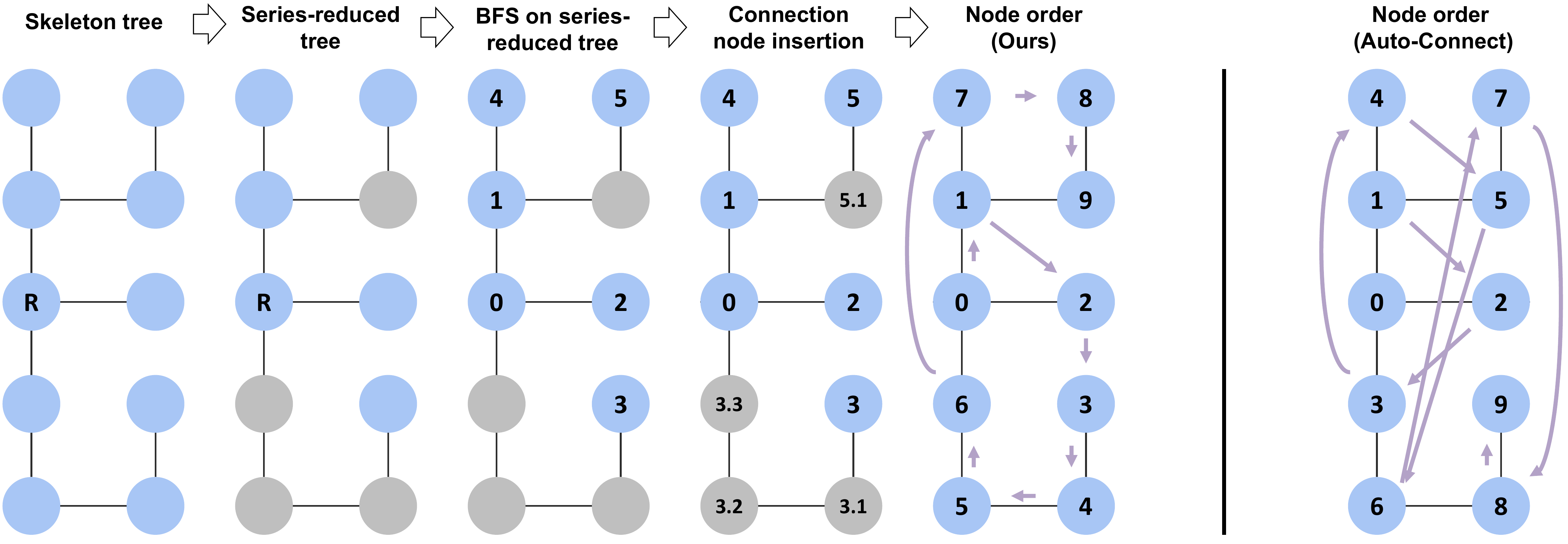}
    \caption{Comparison of node ordering with Auto-Connect. \textbf{Left: illustration of branch-centric tokenization (BCT).} Starting from the input skeleton tree, we first construct its series-reduced tree and run BFS on this reduced tree to order the branches. We then reinsert the smoothed connection nodes along each visited branch, which yields the final node order used by BCT. For clarity, we omit delimiter insertion in this figure. We use \textcolor[HTML]{B3A2C7}{purple} arrows to indicate the node traversal order. \textbf{Right: the node order of Auto-Connect.} As indicated by the arrows, Auto-Connect interleaves nodes from one branch with nodes from other branches in the traversal sequence. In contrast, our branch-level ordering keeps nodes from the same branch adjacent in the serialized sequence.}
    \label{fig:token}
    \vspace{-0.3cm}
\end{figure}

%% file: figures/fig_tokencount.tex
\begin{figure}
        \centering
        \includegraphics[width=0.5\linewidth]{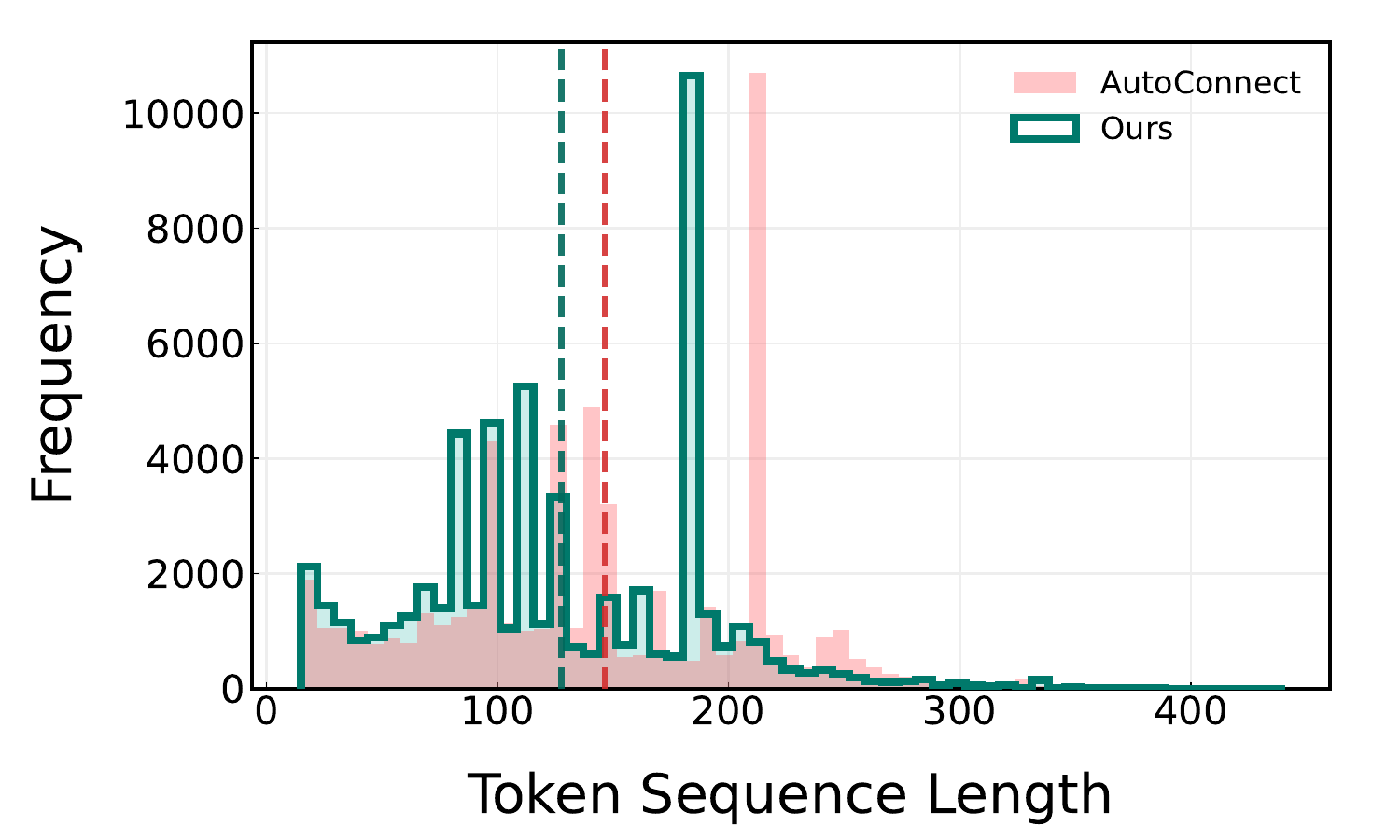}
        \caption{Token sequence length distributions on Articulation-XL2.0. Our branch-centric tokenization on average produces shorter token sequences than Auto-Connect, with mean lengths of $127.6$ vs. $146.3$ tokens (indicated by the dashed lines). Auto-Connect produces longer token sequences than ours in $94.4\%$ of samples.}
        \label{fig:tokencount}
\end{figure}

%% file: sections/tta.tex
\subsection{Test-Time Augmentation for Skeleton Generation}
\label{sec:reranking}
Meshes at test-time can appear in different axis-aligned orientations rather than a single canonical upright frame, as in datasets such as Articulation-XL2.0 and practical 3D asset collections. This creates alignment ambiguity for autoregressive skeleton generation. An intuitive idea is to apply test-time augmentation over multiple rotated views. However, standard test-time augmentation, \eg, for image classification, does not directly apply. Predictions from different views may differ in both topology and joint correspondence; it is \textit{unclear} how to ``aggregate'' these predictions. 

We introduce a coverage-consensus aggregation rule that selects the final prediction using two criteria: mesh coverage and cross-view agreement.

{\bf\noindent View-augmented generation.} We define our augmentation set $\mathcal{A}_r = \{a_1, \dots, a_6\}$ as a discrete set of six rotations, each represented by a rotation matrix $R_k \in \mathrm{SO}(3)$ such that
\[
R_k \mathbf{u}_k = \mathbf{e}_z, \qquad \mathbf{u}_k \in \{\pm \mathbf{e}_x, \pm \mathbf{e}_y, \pm \mathbf{e}_z\},
\]
where $\mathbf{e}_x, \mathbf{e}_y, \mathbf{e}_z$ are the canonical coordinate axes. That is, $\mathcal{A}_r$ contains the identity rotation and five additional shortest-arc rotations that map each axial direction to the canonical $z$-axis. We center this set on the $z$-axis because many 3D assets use $z$ as an upright direction. For an input mesh $\gM$, the model independently predicts a skeleton for each augmented view $a_k(\gM)$. We then map the $k$-th prediction back to the canonical frame by applying the inverse rotation $R_k^{-1}$, yielding the prediction $\widehat{\gT}_k=(\widehat{\gV}_k,\widehat{\gE}_k)$ and its joint set $\widehat{\gJ}_k = \{j^v\}_{v \in \widehat{\gV}_k}$. We denote the set of predictions as $\mathcal{H} = \{\widehat{\gT}_1, \dots, \widehat{\gT}_6\}$ in a shared canonical coordinate system.

{\bf\noindent Coverage-consensus aggregation.}
Given the set of predictions $\mathcal{H}$, we select a single skeleton $\gT^*$ by minimizing a score that considers both coverage and joint consensus:
\bea \label{eq:heuristic_score}
\gT^* = g(\mathcal{H}) \triangleq \argmin_{\widehat{\gT}_k \in \mathcal{H}} \Big(S_{\mathrm{cov}}(\widehat{\gT}_k, \gM) + \lambda_{\tt joint}\cdot S_{\mathrm{joint}}(\widehat{\gT}_k, \mathcal{H}) \Big).
\eea
Observe that $g$ in~\equref{eq:heuristic_score} is invariant to the permutation of the augmentation index $k$ given that the minimum is unique.

In more detail, the coverage score $S_{\mathrm{cov}}$ measures how well a predicted skeleton covers the mesh geometry, with lower values indicating better coverage. Given $M=500$ uniformly sampled surface vertices ${\gV}_{\gM} \subset \gM$, it is defined as:
\bea \label{eq:geom_score}
S_{\mathrm{cov}}(\widehat{\gT}_k, \gM) = \frac{1}{M} \sum_{\mathbf{p} \in {\gV}_{\gM}} -\exp\Big(-\alpha_{r} \min_{e \in \widehat{\gE}_k} \mathcal{D}(\mathbf{p}, e)\Big),
\eea
where $\alpha_r$ dictates the spatial tolerance radius. $\mathcal{D}(\mathbf{p}, e)$ is the orthogonal projection distance from a surface vertex $\mathbf{p}$ to a predicted bone $e=(u,v) \in \widehat{\gE}_k$, which is:
\bea \label{eq:point_to_line}
\mathcal{D}(\mathbf{p}, e) = \min_{\gamma \in [0,1]} \big\| \mathbf{p} - \big(j^u + \gamma(j^v - j^u)\big) \big\|_2.
\eea

The joint-consensus score $S_{\mathrm{joint}}$ measures how well one prediction agrees with the others in joint space. For each hypothesis $\widehat{\gT}_k$, we compute pairwise Sinkhorn distances~\citep{villani2009optimal} between its joint set $\widehat{\gJ}_k$ and the joint sets of the remaining predictions:
\bea
S_{\mathrm{joint}}(\widehat{\gT}_k, \mathcal{H}) = \sum_{\widehat{\gT}_i \in \mathcal{H}, \widehat{\gT}_k\neq\widehat{\gT}_i} \mathcal{D}_{\tt sinkhorn}(\widehat{\gJ}_k, \widehat{\gJ}_i).
\eea
This term favors the predictions that are, on average, closer to the other predictions.

%% file: sections/exp.tex
\section{Experiments}
\label{sec:exp}
In \secref{sec:exp_results}, we compare our method with prior works. We then ablate the tokenization design and the view-augmented generation strategy in~\secref{sec:ablations}. Additional experiments and remaining test-time augmentation ablations are provided in Appendix~\ref{sec:structural}, ~\ref{sec:test-time-ablation}, and~\ref{sec:anymate}. 

\input{tables/tab_main}

\myparagraph{Dataset.} For a fair comparison with previous work~\citep{song2025puppeteer}, we train our model on the Articulation-XL2.0 dataset, which contains $57k$ samples including the diverse-pose training split. For evaluation, we report results on the Articulation-XL2.0 test set, the diverse-pose test split from Articulation-XL2.0, and the ModelsResource test set~\citep{xu2019predicting}.

\myparagraph{Metrics.} Following standard metrics in prior work~\citep{song2025puppeteer}, we report Chamfer Distance between joints (CD-J2J), Chamfer Distance between joints and bones (CD-J2B), and Chamfer Distance between bones (CD-B2B). CD-J2J measures the proximity of predicted joints to the ground truth. CD-B2B and CD-J2B capture both joint-level and bone-level alignment.

\myparagraph{Baselines.} We compare our method against Pinocchio~\citep{baran2007automatic},
RigNet~\citep{xu2020rignet},
UniRig~\citep{zhang2025one},
MagicArticulate~\citep{song2025magicarticulate},
Puppeteer~\citep{song2025puppeteer}, and
Auto-Connect~\citep{guo2025auto}. For completeness, we include the results reported by TokenRig~\citep{zhang2026skin} as a reference. This comparison is not strictly controlled and may favor TokenRig, since TokenRig is trained on Articulation-XL2.0, VRoid Hub, and ModelsResource. In contrast, our model is trained only on Articulation-XL2.0, following the evaluation protocol of RigNet, UniRig, MagicArticulate, and Puppeteer.

\input{figures/fig_qualitative}
\myparagraph{Implementation details.} For the network architecture, we match Auto-Connect by using a Michelangelo-based mesh encoder~\citep{zhao2023michelangelo}, an OPT-style autoregressive decoder~\citep{zhang2022opt}, and level embeddings for each token.
We train our model with Adam for $1000$ epochs, using a base learning rate of $1\times 10^{-4}$ and a one-cycle learning rate schedule~\citep{smith2019super}. The global batch size is $96$. We use $8$-bit quantization, \ie, $n=8$. For training-time data augmentation, we follow previous work~\citep{zhang2025one,guo2025auto} and include random rotation, random translation, random scaling, and bone perturbation. For test-time augmentation, we set $\alpha_r = 15$ and $\lambda_{\tt joint}=1.0$. Training takes approximately $57$ hours on four NVIDIA H100 GPUs. 

\subsection{Results}
\label{sec:exp_results}
\tabref{tab:main} reports quantitative comparisons with prior methods on three test sets. Even without TTA, our model already outperforms all directly comparable baselines/methods. Adding TTA further improves all reported metrics, with consistent gains across both the in-domain Articulation-XL2.0 benchmark and the cross-dataset ModelsResource benchmark. We do not report results for Auto-Connect on the Diverse-pose split because no official pretrained checkpoint is available.

\Figref{fig:qualitative} shows qualitative comparisons with baselines. Our method produces skeletons that better follow the target part layout and local connectivity across diverse shapes, including challenging cases with elongated structures. Comparing our full model with the variant without TTA shows that TTA improves the local structural accuracy of the predictions.

\subsection{Ablations}
\label{sec:ablations}

\myparagraph{Ablation on tokenization.}

\input{tables/tab_ablation_token}
\tabref{tab:ablation_token} isolates the two components of our tokenization design: branch-centric tokenization (BCT) and distance-aware traversal (DAT). Introducing BCT alone already  
 improves all three metrics on both Articulation-XL2.0 and Diverse-pose. This suggests that separating global branch structure from local connection nodes yields a more effective serialization than a naive BFS traversal. Adding DAT further improves performance on top of BCT and gives the best overall results, indicating that a more stable child ordering provides an additional benefit beyond the tokenization structure itself.

\input{tables/tab_ablation_aggregate}
\myparagraph{Ablation on TTA.} We view TTA for autoregressive skeleton generation as a two-stage procedure: candidate generation followed by final selection. Our method implements these stages with view-augmented generation and coverage-consensus aggregation, respectively. We ablate the design choices of the two stages separately. For candidate generation, we compare our six-view rotation set $\mathcal{A}_r$ against beam search under the identity transformation $\mathcal{A}_{\mathrm{id}}$. For the selection, we compare our coverage-consensus aggregation against two common strategies: sequence likelihood and length-normalized sequence likelihood~\citep{wu2016google}. Notably, the combination of beam search and length-normalized likelihood selection corresponds to the inference strategy used in prior work~\citep{zhang2025one}. As shown in \tabref{tab:tta_agg}, coverage-consensus aggregation consistently outperforms likelihood baselines, and combining it with the view augmentation yields the best overall result.

%% file: tables/tab_main.tex
\begin{table*}[t]
\centering
\caption{\textbf{Quantitative comparison with baselines across three test sets.}
We report results on Articulation-XL2.0, ModelsResource, and the diverse-pose split using CD-J2J, CD-J2B, and CD-B2B.
Our method achieves the best results among directly comparable methods. \textbf{Bold} numbers denote the best result among directly comparable methods in each column.
\textcolor{black!55}{*Gray results are reported from TokenRig's paper and are included for reference only, as TokenRig uses additional training data, including the ModelsResource training set.}}
\label{tab:main}

\resizebox{\linewidth}{!}{
\begin{tabular}{l|ccc|ccc|ccc}
\toprule
\multirow{2}{*}{Method}
& \multicolumn{3}{c|}{Articulation-XL2.0} 
& \multicolumn{3}{c|}{ModelsResource} 
& \multicolumn{3}{c}{Diverse-pose} \\
\cline{2-10}
& J2J $\downarrow$ & J2B $\downarrow$ & B2B $\downarrow$
& J2J $\downarrow$ & J2B $\downarrow$ & B2B $\downarrow$
& J2J $\downarrow$ & J2B $\downarrow$ & B2B $\downarrow$ \\
\midrule
Pinocchio~\citep{baran2007automatic} 
& 8.324 & 6.612 & 5.485 
& 6.852 & 4.824 & 4.089 
& 7.967 & 6.411 & 5.149 \\

RigNet~\citep{xu2020rignet}
& 7.618 & 6.076 & 5.279 
& 7.223 & 5.987 & 4.329 
& 7.751 & 6.392 & 5.713 \\

UniRig~\citep{zhang2025one}
& 3.305 & 2.611 & 2.180 
& 3.964 & 3.021 & 2.570 
& 3.252 & 2.569 & 2.077 \\

MagicArti.~\citep{song2025magicarticulate}
& 3.264 & 2.503 & 2.123 
& 4.114 & 3.137 & 2.693 
& 4.376 & 3.456 & 2.955 \\

Puppeteer~\citep{song2025puppeteer}
& {3.109} & {2.370} & {1.983}
& {3.766} & {2.804} & {2.405}
& {2.514} & {1.986} & {1.598} \\

Auto-Connect~\citep{guo2025auto} 
& {2.572} & 2.030 & 1.683
& 3.735 & 2.816 & 2.362
& - & - & - \\

\graycell{TokenRig~\citep{zhang2026skin}*}
& \graycell{2.485} & \graycell{1.599} & \graycell{1.463}
& \graycell{2.893} & \graycell{2.012} & \graycell{1.547}
& \graycell{2.317} & \graycell{1.540} & \graycell{1.446} \\

\hline
Ours w/o TTA
& {2.438} & {1.770} & {1.517}
& 3.547 & 2.632 & 2.115
& {2.088} & {1.557} & {1.338} \\

Ours
& \textbf{2.364} & \textbf{1.687} & \textbf{1.447} 
& \textbf{3.445} & \textbf{2.513} & \textbf{2.016}
& \textbf{{2.052}} & \textbf{{1.521}} & \textbf{{1.296}} \\

\bottomrule
\end{tabular}}
\vspace{-1.mm}
\end{table*}

%% file: figures/fig_qualitative.tex
\begin{figure}[t]
    \centering
    \includegraphics[width=.99\linewidth]{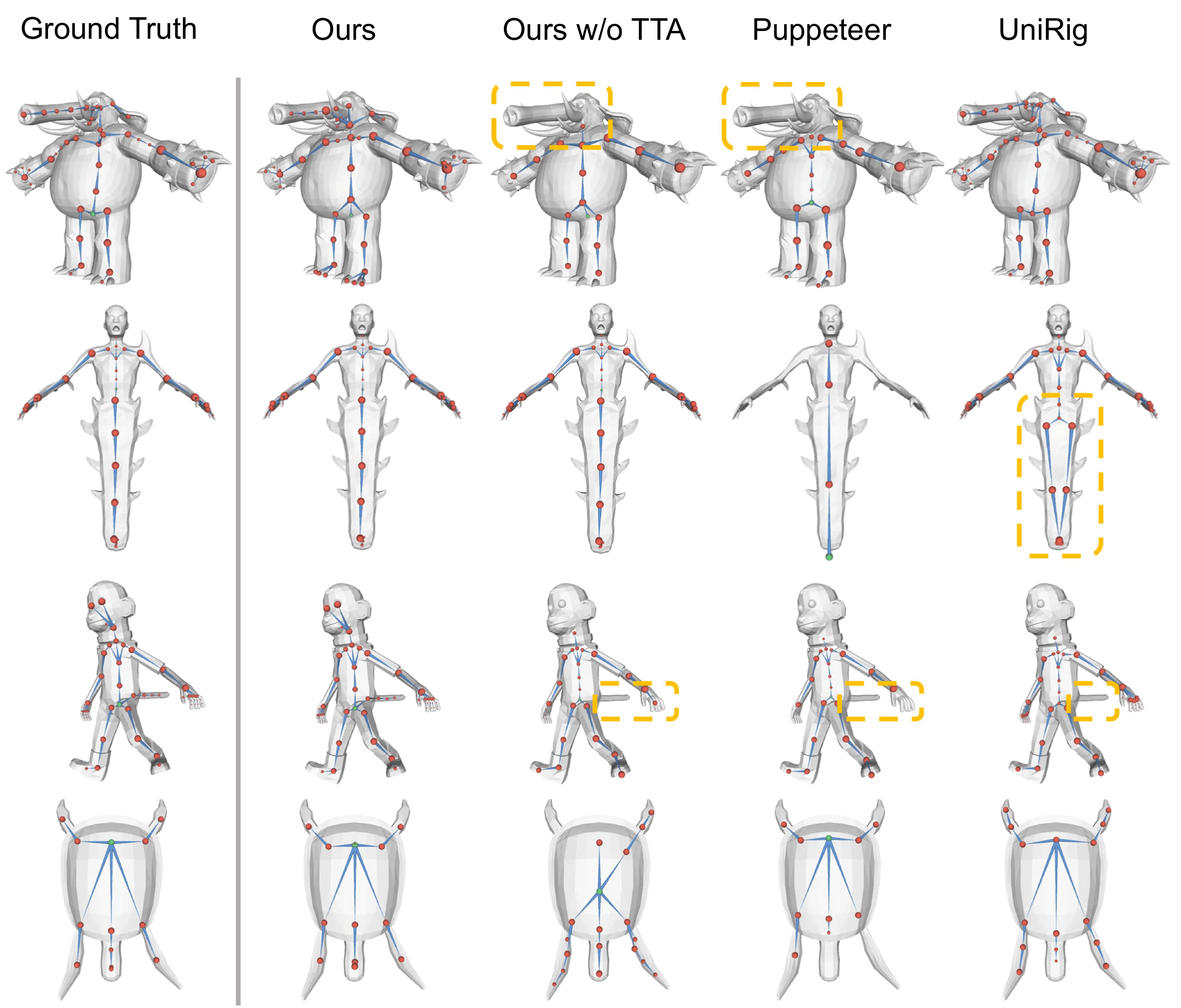}
    \vspace{-0.2cm}
    \caption{{\bf Qualitative comparison on challenging examples from Articulation-XL2.0.} We compare our method, our method without TTA, Puppeteer~\citep{song2025puppeteer}, and UniRig~\citep{zhang2025one} against the ground-truth skeletons. Our method better matches the target connectivity and joint positions, especially around challenging part attachments and elongated structures, while TTA further improves over the variant without TTA. The \textcolor[HTML]{FFC000}{dashed boxes} highlight failure regions.}
    \label{fig:qualitative}
    \vspace{-0.5cm}
\end{figure}

%% file: tables/tab_ablation_token.tex
\begin{table}
\centering
\caption{Ablation on tokenization design choices.}
\label{tab:ablation_token}
{
\begin{tabular}{cc|lll|lll}
\toprule
\multirow{2}{*}{BCT} & \multirow{2}{*}{DAT} 
& \multicolumn{3}{c|}{Articulation-XL2.0} 
& \multicolumn{3}{c}{Diverse-pose} \\ 
\cline{3-8}
&                      
& J2J & J2B & B2B 
& J2J & J2B & B2B \\
\hline
&                      
& 2.543 & 1.864 & 1.582    
& 2.216 & 1.695 & 1.471 \\
$\checkmark$ &                      
& 2.479 & 1.798 & 1.528     
& 2.145 & 1.585 & 1.367 \\
$\checkmark$ & $\checkmark$         
& \textbf{2.438} & \textbf{1.770} & \textbf{1.517} 
& \textbf{2.088} & \textbf{1.557} & \textbf{1.338} \\
\bottomrule
\end{tabular}
}

\end{table}

%% file: tables/tab_ablation_aggregate.tex
\begin{table}[t]
\centering
\caption{Ablation on test-time augmentation on Articulation-XL2.0. We compare two ways to generate candidate skeletons under a fixed six-hypothesis budget: greedy decoding with $\mathcal{A}_r$ (Ours) and beam search under $\mathcal{A}_{\mathrm{id}}$. We then compare different selection rules over the resulting candidates. The best results are shown in \textbf{bold}.}
\label{tab:tta_agg}
{
\begin{tabular}{llll|ccc}
\toprule
Aug. & Sampling & \# Hypo. & Aggregation & J2J $\downarrow$ & J2B $\downarrow$ & B2B $\downarrow$ \\
\midrule
$\mathcal{A}_{\mathrm{id}}$ & Greedy & 1 & - & 2.438 & 1.770 & 1.517 \\
\midrule
$\mathcal{A}_r$ & Greedy & 6 & Likelihood & 2.593 & 1.901 & 1.631 \\
$\mathcal{A}_r$ & Greedy & 6 & Length-normalized likelihood & 2.474 & 1.809 & 1.563 \\
$\mathcal{A}_r$ & Greedy & 6 & Ours & \textbf{2.364} & \textbf{1.687} & \textbf{1.447} \\
\midrule
$\mathcal{A}_{\mathrm{id}}$ & Beam search & 6 & Likelihood & 2.381 & 1.739 & 1.504 \\
$\mathcal{A}_{\mathrm{id}}$ & Beam search & 6 & Length-normalized likelihood & 2.377 & 1.735 & 1.501 \\
$\mathcal{A}_{\mathrm{id}}$ & Beam search & 6 & Ours & 2.375 & 1.735 & 1.501 \\
\bottomrule
\end{tabular}}
\end{table}

%% file: sections/conclusion.tex
\section{Conclusion}
We presented a skeleton generation framework built around two design choices: branch-centric tokenization and view-augmented generation. Branch-centric tokenization reorganizes the serialized skeleton by branch structure and, together with distance-aware traversal, yields a more compact and effective representation than standard BFS-style serialization used in previous work. At inference time, view-augmented generation and coverage-consensus aggregation reduce errors caused by axis alignment ambiguity without changing the training objective or network backbone. Experiments show that these design choices consistently improve both joint accuracy and bone alignment over prior methods. Our ablations further support that the gains come from both the tokenization design and the proposed inference procedure.

%% file: sections/supp.tex
\newpage

\section{Branch-Centric Tokenization Algorithm}
\label{sec:bct_algo}
Algorithm~\ref{alg:branch_tokenization} summarizes BCT. We highlight in green the steps that differ from the tokenization method proposed by Auto-Connect.

\begin{algorithm}[h]
\caption{Branch-Centric Tokenization (BCT)}
\label{alg:branch_tokenization}
\begin{algorithmic}
    \REQUIRE Skeleton tree $\mathcal{T}=(\mathcal{V},\mathcal{E})$ with root $v^{\mathrm{root}}$, quantization bit-width $n$
    \ENSURE Serialized token sequence $\boldsymbol{\tau}$

    \STATE Define $\textsc{AppendCoord}(v)$ to append $(Q_n(j_x^v), Q_n(j_y^v), Q_n(j_z^v))$ to $\boldsymbol{\tau}$
    \STATE \textcolor{ForestGreen}{Construct the series-reduced tree $\mathcal{T}_{\tt sr}$ from \eqref{eq:vback}.}
    \STATE Initialize $\boldsymbol{\tau} \gets [\text{BOS}]$
    \STATE $\textsc{AppendCoord}(v^{\mathrm{root}})$;
    \STATE Append $[\text{E2}]$ to $\boldsymbol{\tau}$
    \STATE $Q \gets \text{Queue}([v^{\mathrm{root}}])$

    \WHILE{$Q$ is not empty}
        \STATE $L \gets \text{length}(Q)$
        \FOR{$i = 1$ \textbf{to} $L$}
            \STATE $p \gets Q.\text{dequeue}()$
            \STATE \textcolor{ForestGreen}{Let $\mathcal{V}_{\tt child}$ be the children of $p$ in $\mathcal{T}_{\tt sr}$, sorted by DAT in ascending Euclidean distance to $p$}
            \FORALL{$v \in \mathcal{V}_{\tt child}(p)$}
                \STATE $\textsc{AppendCoord}(v)$
                \STATE \textcolor{ForestGreen}{Let $\mathcal{S}_{v \to p}$ be the connection nodes on the path from $v$ to $p$ in $\mathcal{T}$, ordered from $v$ toward $p$}
                \FORALL{$c \in \mathcal{S}_{v \to p}$}
                    \STATE \textcolor{ForestGreen}{$\textsc{AppendCoord}(c)$}
                \ENDFOR
                \STATE \textcolor{ForestGreen}{Append $[\text{E3}]$ to $\boldsymbol{\tau}$}
                \STATE $Q.\text{push}(v)$
            \ENDFOR
            \STATE Append $[\text{E1}]$ to $\boldsymbol{\tau}$
        \ENDFOR
        \STATE Append $[\text{E2}]$ to $\boldsymbol{\tau}$
    \ENDWHILE

    \STATE Append $[\text{EOS}]$ to $\boldsymbol{\tau}$
    \STATE \textbf{return} $\boldsymbol{\tau}$
\end{algorithmic}
\end{algorithm}

\section{Evaluation with Structural Metrics}
\label{sec:structural}
\input{tables/tab_topology}
We further evaluate the structural similarity of the predicted skeletons using Graph2Vec distance~\citep{narayanan2017graph2vec} and tree edit distance (TED). We train the Graph2Vec model on skeletons from the Articulation-XL2.0 training set and compute the cosine distance between the embeddings of each predicted skeleton and its ground-truth counterpart. For TED, we order the children of each node by the size of their corresponding subtrees and compute the distance using the Zhang--Shasha algorithm~\citep{zhang1989simple}. As shown in \tabref{tab:ted_graph2vec}, our model achieves the lowest TED and Graph2Vec distance among all compared methods.
\input{sections/supp_tta}

\section{Additional Experiment on the Anymate Dataset}
\label{sec:anymate}
\input{tables/tab_anymate}
We additionally train our model on the Anymate dataset~\citep{deng2025anymate}. Following Auto-Connect, we assume that the skeleton forms a singly connected tree, and therefore exclude training samples with self-loops or multiple connected components, resulting in a clean training set of $205k$ samples. For evaluation, we use the original Anymate test set to ensure a fair comparison. Following the original paper, we report Chamfer Distance (CD) and Earth Mover's Distance (EMD), where EMD measures the optimal transport cost between predicted joints and ground-truth joints. We compare against Pinocchio~\citep{baran2007automatic}, RigNet~\citep{xu2020rignet}, Anymate~\citep{deng2025anymate}, and UniRig~\citep{zhang2025one}. Among them, Anymate directly predicts joint positions without autoregressive generation. UniRig is retrained on Anymate using the official codebase for a fair comparison.

For our method, all hyperparameters are the same as in the main text, except that we train for only $300$ epochs because of the larger dataset size. Training takes approximately $61$ hours. We do not directly evaluate the model trained on Articulation-XL2.0, because Anymate and Articulation-XL2.0 are both sourced from Objaverse-XL and may therefore have test-set overlap. As shown in \tabref{tab:anymate}, our method achieves the lowest CD and EMD among the compared methods.

\section{Inference Time Comparison}
\label{sec:supp_inference_time}
\input{tables/tab_inference_time}

In \tabref{tab:inference_time}, we compare inference time with prior autoregressive methods. Our model without test-time augmentation (TTA) is faster than UniRig, Puppeteer, and TokenRig. With TTA, inference time increases as expected because we decode six rotated views, but the overall runtime remains moderate. For all baselines, we use the official code, exclude data processing time, and run inference on a single NVIDIA H100 GPU.

\section{Limitations}
\label{sec:supp_limitations}

Our method has two main limitations. First, our view augmentation design is motivated by the fact that meshes are not consistently canonicalized to a shared upright orientation. We follow the evaluation protocol of prior baselines, which operate on meshes in their original orientations. Stroke3D~\citep{zhao2026stroke3d} highlighted this issue and attempted to address it by aligning meshes to a consistent upright direction. However, its upright-detection system is not publicly available, so we could not evaluate our method and prior baselines under a setting where all meshes are canonicalized to a consistent $z$-upright orientation. If a reliable orientation-canonicalization method becomes available in the future, the need for the view augmentation component of our TTA design may be reduced. That said, \tabref{tab:stochastic_sampling} shows that our coverage-consensus aggregation remains effective even when the hypotheses are not generated by view augmentation, which suggests that it is still useful beyond this particular assumption.

Second, we do not study post-training in this paper. Prior work, including TokenRig~\citep{zhang2026skin} and Auto-Connect~\citep{guo2025auto}, has shown that post-training can be effective for improving skeleton generation. We intentionally exclude it here because our goal is to study the effects of branch-centric tokenization and test-time augmentation in a clean, controlled setting. An important future direction is to combine our method with post-training and test whether the gains are complementary.

%% file: tables/tab_topology.tex
\begin{table*}[h]
\centering
\caption{\textbf{Quantitative comparison of skeleton prediction methods on Articulation-XL2.0.}
We report TED and Graph2Vec distance on the Articulation-XL2.0 test set. \textbf{Bold} numbers denote the best result in each column.}
\label{tab:ted_graph2vec}

{
\begin{tabular}{l|cc}
\toprule
Method
& TED $\downarrow$
& Graph2Vec distance $\downarrow$ \\
\midrule

MagicArti.
& 6.837
& 0.254 \\

Puppeteer
& 6.736
& 0.245 \\

TokenRig
& 7.120
& 0.235 \\

\hline

Ours
& \textbf{5.984}
& \textbf{0.222} \\

\bottomrule
\end{tabular}}
\vspace{-1.mm}
\end{table*}

%% file: sections/supp_tta.tex
\section{Test-Time Augmentation Ablations}
\label{sec:test-time-ablation}
We provide additional ablations of the test-time augmentation strategy used in our method. We focus on four questions. First, does a larger axis-aligned rotation group help? Second, is coverage-consensus aggregation more effective than likelihood baselines under common stochastic sampling? Third, which augmentation choices are effective? Finally, how sensitive is coverage-consensus aggregation to its hyperparameters?

\subsection{Alternative Rotation Set}
We evaluate a larger 24-element cube rotation group $\mathcal{A}_{\mathrm{cube}}$, which enumerates all right-handed axis-aligned rotations~\citep{gallian2021contemporary} for view-augmented generation. Compared with the six-view set $\mathcal{A}_r$ used in the main paper, this larger set does not rely on prior knowledge of common upright orientations.

\input{tables/tab_ablation_cube}

\paragraph{Results.}
\tabref{tab:tta_cube} shows that expanding the rotation set from $\mathcal{A}_r$ to $\mathcal{A}_{\mathrm{cube}}$ further improves performance, but only modestly, while requiring four times as many decoded hypotheses. We therefore use the smaller set $\mathcal{A}_r$ in the main method because it provides a better accuracy-efficiency trade-off.

\subsection{Alternative Sampling Method}
\input{tables/tab_stochastic_sampling}
To supplement \tabref{tab:tta_agg}, we study a second common way to generate candidate skeletons: stochastic sampling under the identity transformation $\mathcal{A}_{\mathrm{id}}$. We then compare coverage-consensus aggregation against standard likelihood-based selection rules on the resulting samples. As shown in \tabref{tab:stochastic_sampling}, coverage-consensus aggregation outperforms both likelihood-based baselines in this setting. For reference, the table also reports view-augmented greedy decoding under $\mathcal{A}_r$, which achieves the best overall performance.

\subsection{Augmentation Choice}
We next fix the decoding strategy to greedy decoding and study which augmentations are useful for view-augmented generation. Throughout this experiment, final selection is performed using our coverage-consensus aggregation.

\paragraph{Augmentations.}
We compare three augmentation choices: the six-view rotation set $\mathcal{A}_r$ defined in \secref{sec:reranking}, random translation jitter, and random scaling. We evaluate each augmentation individually and apply it globally to both the mesh and the skeleton to preserve equivariance. For random translation jitter, we first normalize each sample into $[-1.0,1.0]^3$ and apply a deterministic margin scale of $0.9$. We then sample an axis-wise shift from $[-0.1,0.1]$ so that the translated sample remains within the valid spatial range. For random scaling, we sample an isotropic scale factor from $[0.5,1.0]$.

\paragraph{Results.}
Table~\ref{tab:tta_aug_supp} shows that under coverage-consensus aggregation, the rotation set $\mathcal{A}_r$ is the strongest option. In contrast, random translation jitter and random scaling substantially degrade performance, with random scaling causing the largest drop.

\input{tables/tab_ablation_aug}

\subsection{Hyperparameter Sensitivity of Coverage-Consensus Aggregation}
We finally study the sensitivity of coverage-consensus aggregation to its two main hyperparameters: the spatial tolerance radius $\alpha_r$ and the joint-consensus weight $\lambda_{\tt joint}$.

\input{tables/tab_tta_hyper}

\paragraph{Results.}
Table~\ref{tab:heuristic_hyperparams} shows that the joint-consensus term is important. Using only the coverage term gives consistently worse results across all values of $\alpha_r$. In contrast, introducing a nonzero joint-consensus weight yields a clear improvement. Once $\lambda_{\tt joint}$ is set to a moderate value, performance is fairly stable. We use $\alpha_r=15.0$ and $\lambda_{\tt joint}=1.0$ in the main experiments.

%% file: tables/tab_ablation_cube.tex
\begin{table}[t]
\centering
\caption{Ablation on a larger rotation group for view-augmented generation on Articulation-XL2.0. We compare the six-view rotation set $\mathcal{A}_r$ used in the main paper with the larger 24-view cube rotation set $\mathcal{A}_{\mathrm{cube}}$. The larger set gives a small accuracy gain at the cost of four times more decoded hypotheses.}
\label{tab:tta_cube}
\resizebox{0.6\linewidth}{!}{
\begin{tabular}{lcc|ccc}
\toprule
Aug. & Sampling & \# Hypo. & J2J $\downarrow$ & J2B $\downarrow$ & B2B $\downarrow$ \\
\midrule
$\mathcal{A}_r$ & Greedy & 6 & 2.364 & 1.687 & 1.447 \\
$\mathcal{A}_{\mathrm{cube}}$ & Greedy & 24 & \textbf{2.339} & \textbf{1.673} & \textbf{1.438} \\

\bottomrule
\end{tabular}}
\end{table}

%% file: tables/tab_stochastic_sampling.tex
\begin{table}[t]
\centering
\caption{Comparison of selection strategies for stochastic samples without view-augmented generation on Articulation-XL2.0. The first row reports greedy decoding under $\mathcal{A}_r$ for reference. For stochastic samples under $\mathcal{A}_{\mathrm{id}}$, coverage-consensus aggregation outperforms likelihood-based selection.}
\label{tab:stochastic_sampling}
{
\begin{tabular}{llll|ccc}
\toprule
Aug. & Sampling & \# Hypo. & Aggregation & J2J $\downarrow$ & J2B $\downarrow$ & B2B $\downarrow$ \\
\midrule
$\mathcal{A}_r$ & Greedy & 6 & Coverage-consensus aggregation & \textbf{2.364} & \textbf{1.687} & \textbf{1.447} \\
\midrule
$\mathcal{A}_{\mathrm{id}}$ & Stochastic & 6 & Likelihood & 2.550 & 1.884 & 1.612 \\
$\mathcal{A}_{\mathrm{id}}$ & Stochastic & 6 & Length-normalized likelihood & 2.441 & 1.779 & 1.524 \\
$\mathcal{A}_{\mathrm{id}}$ & Stochastic & 6 & Coverage-consensus aggregation & 2.408 & 1.734 & 1.487 \\

\bottomrule
\end{tabular}}
\vspace{-3mm}
\end{table}

%% file: tables/tab_ablation_aug.tex
\begin{table}[t]
\centering
\caption{Additional ablation on test-time augmentation. Each augmentation is evaluated individually using our coverage-consensus aggregation.}
\label{tab:tta_aug_supp}
{
\begin{tabular}{l|ccc}
\toprule
Augmentation & J2J $\downarrow$ & J2B $\downarrow$ & B2B $\downarrow$ \\
\midrule
Rotation set ($\mathcal{A}_r$) & \textbf{2.364} & \textbf{1.687} & \textbf{1.447} \\
Random translation & 4.752 & 3.128 & 2.202 \\
Random scaling & 8.291 & 5.324 & 3.113 \\
\bottomrule
\end{tabular}}
\end{table}

%% file: tables/tab_tta_hyper.tex
\begin{table}[t]
\centering
\caption{Hyperparameter sensitivity of coverage-consensus aggregation on Articulation-XL2.0. We report CD-J2J over the spatial tolerance radius $\alpha_r$ and the joint-consensus weight $\lambda_{\tt joint}$ under the 6-hypothesis rotation TTA setting with $\mathcal{A}_r$. Bold indicates the best result.}
\label{tab:heuristic_hyperparams}
{
\begin{tabular}{c|cccccc}
\toprule
\multirow{2}{*}{$\alpha_r$} & \multicolumn{6}{c}{Joint-consensus weight $\lambda_{\tt joint}$} \\
\cmidrule(lr){2-7}
 & 0.00 & 0.25 & 0.50 & 1.00 & 2.00 & 4.00 \\
\midrule
5.0  & 2.516 & 2.402 & 2.395 & 2.366 & {2.364} & 2.381 \\
10.0 & 2.481 & 2.393 & 2.377 & {2.364} & 2.367 & 2.381 \\
15.0 & 2.478 & 2.378 & 2.371 & {2.364} & 2.368 & 2.385 \\
20.0 & 2.472 & 2.377 & 2.372 & \textbf{2.363} & 2.374 & 2.383 \\
25.0 & 2.469 & 2.375 & 2.374 & {2.364} & 2.377 & 2.387 \\
\bottomrule
\end{tabular}}
\end{table}

%% file: tables/tab_anymate.tex
\begin{table}[t]
\centering
\caption{Quantitative results on the Anymate dataset~\citep{deng2025anymate}. Our method achieves the best CD and EMD.}
\label{tab:anymate}
\begin{tabular}{l|cc}
\toprule
Method  & CD $\downarrow$ & EMD $\downarrow$ \\ \hline
Pinocchio~\citep{baran2007automatic} & 0.198 & 0.659 \\
RigNet~\citep{xu2020rignet}          & 0.089 & 0.127 \\
Anymate~\citep{deng2025anymate}      & 0.077 & 0.098 \\
UniRig~\citep{zhang2025one}          & 0.096 & 0.135 \\
Ours         & \textbf{0.038}     & \textbf{0.046}    \\
\bottomrule
\end{tabular}
\end{table}

%% file: tables/tab_inference_time.tex
\begin{table}[t]
\centering
\caption{Inference time comparison on Articulation-XL2.0 test set. Measured in seconds.}
\label{tab:inference_time}
\begin{tabular}{lccccc}
\toprule
Method & UniRig  & Puppeteer & TokenRig & Ours w/o TTA & Ours \\
\midrule
Inference time & 1.8 & {1.3} & 7.5 & \textbf{1.0} & 5.8 \\
\bottomrule
\end{tabular}
\end{table}